\documentclass[conference]{IEEEtran}
\IEEEoverridecommandlockouts
\usepackage{amsmath,amssymb,amsfonts}
\usepackage{algorithmic}
\usepackage{graphicx}
\usepackage{textcomp}
\usepackage{xcolor}
\usepackage{lipsum}
\usepackage{lipsum}
\usepackage{graphicx}
\usepackage{pdflscape}
\usepackage{tabularx}
\usepackage{adjustbox}
\usepackage{biblatex}
\usepackage{multirow}
\usepackage{url}
\def\BibTeX{{\rm B\kern-.05em{\sc i\kern-.025em b}\kern-.08em
   T\kern-.1667em\lower.7ex\hbox{E}\kern-.125emX}}
\begin{document}

\title{Peduncle Gripping and Cutting Force for Strawberry Harvesting Robotic End-effector Design
\\
\thanks{This work was supported by Lincoln Agri-Robotics as part of the Expanding Excellence in England (E3) Programme.}
}

\author{\IEEEauthorblockN{Vishnu Rajendran S$^{1, 3}$, Soran Parsa$^{2, 3}$, Simon Parsons$^{2, 3}$, Amir Ghalamzan Esfahani$^{1, 3}$}
\IEEEauthorblockA{\textit{Lincoln Institute of Agri-Technology (LIAT)$^{1}$, School of Computer Science$^{2}$, Lincoln Center of Autonomous Systems (LCAS)$^3$} \\
\textit{University of Lincoln}\\
Lincoln, United Kingdom \\
%%Draft - Version 4.0
}
}

\maketitle

\begin{abstract}
Robotic harvesting of strawberries has gained much interest in the recent past. Although there are many innovations, they haven't yet reached a level that is comparable to an expert human picker. The end effector unit plays a major role in defining the efficiency of such a robotic harvesting system. Even though there are reports on various end effectors for strawberry harvesting, but there they lack a picture of certain parameters that the researchers can rely upon to develop new end effectors. These parameters include the limit of gripping force that can be applied on the peduncle for effective gripping, the force required to cut the strawberry peduncle, etc. These estimations would be helpful in the design cycle of the end effectors that target to grip and cut the strawberry peduncle during the harvesting action. This paper studies the estimation and analysis of these parameters experimentally. It has been estimated that the peduncle gripping force can be limited to 10 N. This enables an end effector to grip a strawberry of mass up to 50 grams with a manipulation acceleration of 50 m/s$^2$ without squeezing the peduncle. The study on peduncle cutting force reveals that a force of 15 N is sufficient to cut strawberry peduncle using a blade with a wedge angle of 16.6$^0$ at 30$^0$ orientation.

\end{abstract}


\section{Introduction}

%Strawberries are among the high-value crops becoming one of daily must-be-used across different countries. The fresh strawberry market The global Fresh Strawberry market was valued at 1855.28 Million USD in 2020 and will grow with a CAGR of 3.35\% from 2020 to 2027~\cite{fsmarket2020}.
%It is mainly driven by its wide incorporation in various processed foods such as shortcake and pastries. Fresh strawberry is rich in vitamins, fibres, and antioxidants due to which it is also utilized in many cosmetics and personal care products. Fresh strawberries are a sodium-free, fat-free, cholesterol-free, low-calorie food and is considered to be a potent fruit in lowering cholesterol and blood pressure. Italy, Germany and the U.K are among the highest strawberry consuming countries in the region~\cite{strawberryconsumption}.

Strawberries are one among the high-yield crops that have received great attention from researchers in developing robotic solutions for harvesting. Robotic harvesting of strawberries has been predominantly researched and developed for those which are grown in greenhouse scenarios\cite{defterli2016review}. A selective harvesting method is primarily utilized to harvest the strawberries because all fruits on the plant may not be ripe at the same time. And moreover, it needs some judgment to pick the ripe fruit which may be staying in isolated condition or as partially/completely occluded by adjoining fruits or by the parent plant. The clustered nature of strawberries makes it troublesome for any robotic intervention to reach an efficiency competent to a human picker. Several strawberry harvesting robots are reported in the literature~\cite{hayashi2010evaluation,han2012strawberry,feng2012new,arima2004strawberry,xiong2019development,yamamoto2014development,cui2013study,kondo1998strawberry}.% and presented by companies\cite{agrobot,rubion,dogtooth,tortuga}.

%\begin{figure}[tb!]
%\centerline{\includegraphics[scale=0.43]{robotex.JPG}}
%\caption{Examples of strawberry harvesting robots: Agrobot\cite{agrobot}, Dogtooth\cite{dogtooth}, Rubicon by Octinion\cite{rubion} (from left to right)}
%\label{robotex}
%\end{figure}

%Bac et al.~\cite{bac2014harvesting,defterli2016review} reviewed related robotic technologies. 
In these robots, the end effector (EE) is the unit that directly and physically interacts with crops. Across different EE technologies for strawberry harvesting, these physical interactions include: (i) gripping/grasping the strawberry by its peduncle or fruit body (attachment), (ii) detaching from the plant by pulling, twisting the fruit, or cutting the peduncle, (iii) facilitating the fruit transport from the detachment location to the storage, and (iv) pushing/parting of strawberries in the cluster for obstacle separation during detaching action \cite{xiong2020obstacle} which is a recently explored functionality.  

\begin{figure}[bt!]
\centerline{\includegraphics[scale=0.48]{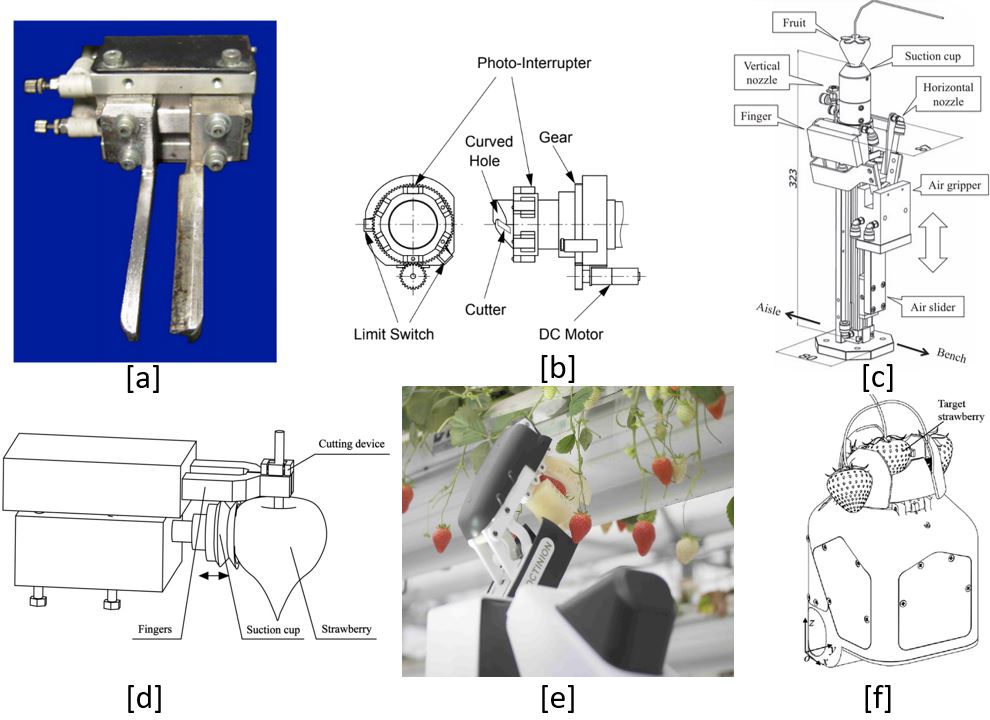}}
\caption{Strawberry harvesting end effectors: [a]. Parallel jaw mechanism for simultaneous attachment and detachment\cite{hayashi2010evaluation}, [b].Suction head for attachment and blade action for detachment\cite{arima2004strawberry}, [c]. Suction head and two jaw fingers for attachment and bending action of end effector for detachment\cite{yamamoto2014development}, [d].Suction cup for attachment and thermal based detachment\cite{feng2012new}, [e]. Soft gripping for attachment and followed by rotation with pulling action for detachment\cite{de2018development}, [f]. Detachment without any attachment\cite{xiong2018design}}
\label{EE}
\end{figure}

Among attachment and detachment actions, some EEs performs simultaneous gripping and detachment of the strawberry peduncle using a parallel jaw mechanism \cite{hayashi2010evaluation,hayashi2014field} (figure \ref{EE}a). In such a mechanism, one jaw will be shaped in the form of a cutting blade or is provided with a provision to attach replaceable blades. One such EE design makes use of a suction cup to provide an additional grip by sucking the fruit body to avoid any positional errors during this simultaneous gripping and cutting actions~\cite{hayashi2010evaluation}. A suction-based approach is used by another end effector which uses a suction head to grip the fruit body and then rotates so that a blade arranged on the curved opening of the suction head trims the peduncle~\cite{arima2004strawberry}~(figure~\ref{EE}b). Instead of using any blades for cutting the peduncle, the EE reported by Yamamoto S et al.~\cite{yamamoto2014development} uses a bending action to detach the strawberry after gripping the fruit body by a suction head and a two-jaw gripper~(figure \ref{EE}c). Also, a thermal-based cutting is reported for another EE that uses an electrically heated wire on the gripping jaw to cut the peduncle~\cite{feng2012new}. In this EE, once the fruit body is gripped by the suction cup to position the peduncle between the two jaws (cutting device), the jaws then close and cuts the peduncle using the heated wire~(figure \ref{EE}d). The end effector developed by  Octinion uses a soft gripper to grip the strawberry fruit body and imposes a rotational motion while pulling the strawberry to detach it from the peduncle~\cite{de2018development}~(figure \ref{EE}e). The EEs reported above make either a gripping contact with the fruit body or with the peduncle during the harvesting action. But the EE reported by Xiong Y et al.~\cite{xiong2018design} doesn't grip the fruit body or the peduncle during the harvesting action. Instead, a combination of three active and passive fingers guides the strawberry into the EE housing. Once the fruit reaches the cutting location, scissor-shaped blades cut the peduncle to detach the strawberry~(figure \ref{EE}f).

Considering the different options for gripping and cutting, it is always beneficial to avoid applying any forces on the fruit body by the EE contact surfaces. Since strawberry is a very soft and delicate fruit, there are higher chances of bruising during such operations. Aliasgarian et al.~\cite{aliasgarian2013mechanical} showed that the strawberry fruits are more damaged when exposed to compression forces on their body. Hence, from the EE design point of view, it is recommended to target the peduncle for gripping/cutting actions or to avoid any grip action as demonstrated by Xiong Y et al.~{\cite{xiong2018design}}.

%To develop an end effector solution that attempts to detach strawberries by targeting the peduncle, it is essential to understand certain physical properties of the peduncle. This includes the estimate of the required cutting force and the gripping force that can be applied on the peduncle. Knowing  the cutting force while using a particular cutting blade profile, gives insight for a better selection of actuation system to provide the required cutting force. Moreover, the practice of using off-the-shelf blades takes away the need of investing effort to design optimum blade profiles. Rather such blades can be directly used or can be custom-made according to the standard profile. If the end effector is designed in such a way to use interchangeable blades, replacing the blades would be easier during worn-out situations. Hence it is wise to use cutting blades with a standard profile and in an interchangeable configuration in the end effectors.\\
To develop an end effector solution that attempts to detach strawberries by cutting the peduncle, it is essential to understand certain design parameters. This includes the estimation of the required cutting and the gripping force that can be applied on the peduncle. Knowing the limit of gripping force eliminates the chances of a loose grip or crushing the peduncle during the gripping action. These situations can lead the detached strawberry to fall off from the grip while manipulating it during the harvesting cycle. Moreover, knowing these force gives an insight for the selection of actuation system for the EE. %Also, it can be used as a reference when tactile-based sensors are used to control the gripping force. \\

The only study that we are aware of which reports the measurement of the force required to detach the peduncle from a strawberry is by Dimeas F et al.~{\cite{dimeas2013towards}}. This considers the use of a pulling effort (technique A) and a combination bending-pulling effort (technique B). Technique A needed a force of 13.94 N for a peduncle with  mean diameter of 2.17 mm, while technique B took 3.17 N to detach a peduncle with a mean diameter of 1.78 mm. Moreover, Dimeas F et al.~{\cite{dimeas2013towards}} also reported that a safe gripping force below 10 N can be applied on the strawberry fruit body for detaching the peduncle by technique B such that it won't bruise the fruit body. Apart from these measured forces, no work has been reported which studied the cutting force and the limit of gripping force that can be applied on the strawberry peduncle. Given this lack of data, this paper studies, the cutting and gripping force that must applied to the strawberry peduncle, helping future work on EEs targeting peduncle for the selective harvesting of strawberries. 

%Along with these studies, the authors have also made field studies to estimate certain physiological features of strawberries, i.e., the mass of fully ripe strawberries and their respective peduncle diameter. The mass details can help researchers in getting an overview of the approximate mass of a ripe strawberry that needs to be handled by the EE. Moreover, the peduncle diameter range helps decide the baseline opening distance of gripping fingers while approaching a strawberry peduncle. Since the end effectors have to target strawberries from clusters as well, the controlled opening of fingers would be required.

For presenting the study details, this paper is organized in the following order. Section II outline the experimental studies conducted, and section III discusses the results obtained. The conclusive remarks of the paper are presented in section IV followed by study limitations in section V.

%\section{Field Study}

%This field study has been made to estimate the mass of fully ripe strawberries and their respective peduncle diameter. The study is made in a polytunnel at Riseholme Campus of the University of Lincoln, UK, where two varieties of strawberries are grown, namely Zara and Katrina. During the study, 100 ripe strawberries of both varieties were harvested leaving 10-15 mm of peduncle on the strawberry. Hence, the measured mass (M$_E$) includes the mass of the strawberry and the mass of the leftover peduncle. If a robot picks a strawberry by its peduncle, the mass handled will involve the mass of the strawberry and this leftover peduncle. This mass was measured using a calibrated weight gauge and was recorded against the diameter of the leftover peduncle (at the trimmed end) measured using a digital vernier caliper. The range of strawberry mass and peduncle diameter, and the variation of the mass (M$_E$) with the respective peduncle diameter are presented in section IV.
%\begin{figure}[tb!]
%\centerline{\includegraphics[scale=0.38]{polytunnel.PNG}}
%\caption{Polytunnel at Riseholme Campus of the University of Lincoln, UK}
%\label{polytunnel}
%\end{figure}
\section{Experimental Study}
For these experimental studies, we have used two varieties of strawberries namely, Katrina and Zara grown in the polytunnel at Riseholme Campus of the University of Lincoln, UK. The peduncle specimens were collected from both varieties in such a way that there will be uniformity in their diameter ranges for better results comparison.

\subsection{Study on the gripping force limit}

This study intended to estimate the limit of the gripping force that can be applied on the strawberry peduncle without crushing it. To understand this force limit, experiments were conducted by applying compression force (analogous to the gripping force) to the peduncle specimens using a Universal Testing Machine (UTM). The peduncles of ripe strawberries of both varieties were selected for preparing the specimens. 15 specimens of each variety were prepared so that the peduncle was 10 mm in length and were trimmed at a distance of 10 mm from the top surface of the ripe strawberry fruit. This specimen measurement can simulate the situation where a two jaw parallel end effector grips the peduncle within 10-20 mm from the top surface of the strawberry top surface during harvesting. The specimen diameter varied from 1.40 mm to 2.22 mm for Katrina with a mean and standard deviation of 1.75 mm and 0.24 mm. And for Zara, the diameter varied from 1.43 mm to 2.33 mm with a mean and standard deviation of 1.76 mm and 0.25mm. 

\begin{figure}[tb!]
\centerline{\includegraphics[scale=0.5]{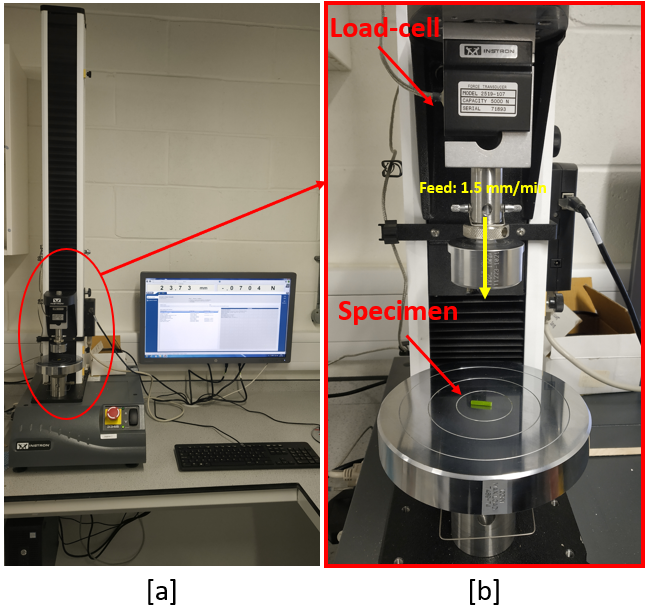}}
\caption{Universal testing machine (UTM) for applying and recording the compression force on specimens used for our experimental study. }
\label{fig1}
\end{figure}

During the experiment, each specimen was placed on the base plate of UTM as shown in the figure \ref{fig1}. The UTM was configured in such a way that the vertical ram (with load cell attachment) approached the base plate at a rate of 1.5 mm/min. Once the ram starts touching the specimen, the connected PC records the interaction force at a resolution of 0.0001 N. The force values recorded during the study are presented in section~\ref{sec:results}. 

\subsection{Study on cutting force}

\begin{figure}[tb!]
\centerline{\includegraphics[scale=0.32]{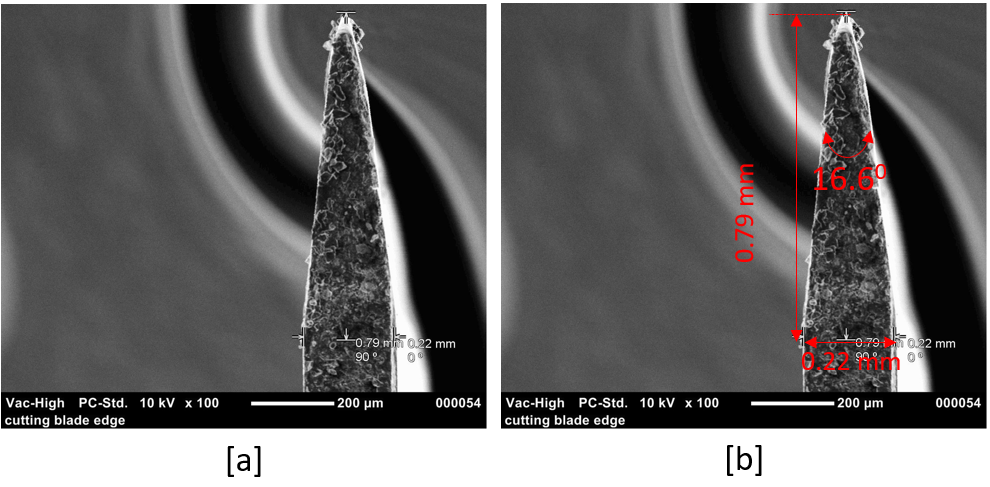}}
\caption{Blade profile measured using Scanning Electron Microscope}
\label{fig2}
\end{figure}

This study aimed to estimate the force required to cut the peduncle of a ripe strawberry using an off-the-shelf blade profile. As an extension to this, the variation of this force at different cutting orientations was also studied. The profile of the selected cutting blade was studied using a scanning electron microscope. The blade had a double bevel cutting edge with a wedge angle of 16.6$^0$ and a thickness of 0.22 mm (figure \ref{fig2}).

\begin{figure}[tb!]
\centerline{\includegraphics[scale=0.55]{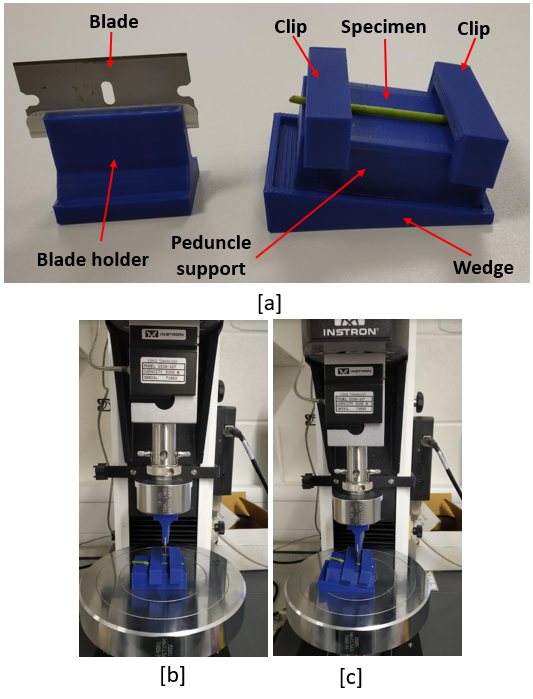}}
\caption{(a). 3d printed fixtures (b). Experimental setup of studying cutting forces and its variation with change in angle of cut}
\label{fig3}
\end{figure}

As shown in the figure \ref{fig3}a, we designed and 3D printed some fixtures in order to conduct the experiment. These include a blade holder for holding the blade, a peduncle support on which to place the peduncle specimen, and a pair of press-fit clips for positioning the specimen on the support. Also, angular wedges of 10$^0$, 20$^0$, 30$^0$ were made that can be used to tilt the peduncle support loaded with the specimen. This allowed us to study the cutting force variation for 0$^0$,10$^0$, 20$^0$, 30$^0$ orientations. The same UTM used in the previous study was used here as well. The blade holder was attached to the vertical ram of UTM using a thin layer of adhesive and the peduncle support to the base plate as shown in the figure \ref{fig3}b-c. 15 peduncle samples from both Zara and Katrina varieties were prepared for this study, i.e., 15 samples from each variety for each orientation of cut. These specimens were prepared by cutting the peduncle at a length 30 mm from the top surface of the ripe strawberry.The diameter range of the specimens used for each trails are shown in the table \ref{table 3}.

\begin{table}[]
\caption{Diameter of the peduncle specimens used to study the cutting force}
\centering
\label{table 3}
\begin{tabular}{|c|c|c|c|c|}
\hline
\multicolumn{5}{|c|}{\textbf{Specimen Diameter (mm)}}                                                                                                                                                 \\ \hline
\multirow{2}{*}{\textbf{Trials}} & \multicolumn{2}{c|}{\textbf{Zara}}                                            & \multicolumn{2}{c|}{\textbf{Katrina}}                                         \\ \cline{2-5} 
                                 & \multicolumn{1}{l|}{\textbf{Minimum}} & \multicolumn{1}{l|}{\textbf{Maximum}} & \multicolumn{1}{l|}{\textbf{Minimum}} & \multicolumn{1}{l|}{\textbf{Maximum}} \\ \hline
\multirow{2}{*}{\textbf{0$^0$}}      & 1.03                                     & 2.14                                    & 1.10                        & 2.00                                   \\ \cline{2-5} 
             & \multicolumn{2}{c|}{Mean(SD):1.53(0.31)}                                                 & \multicolumn{2}{c|}{Mean(SD): 1.51(0.29)}                                           \\ \hline
\multirow{2}{*}{\textbf{10$^0$}}     & 1.20                    & 2.10                            & 1.00                                     & 2.00                                     \\ \cline{2-5} 
                                 & \multicolumn{2}{c|}{Mean(SD): 1.59(0.30)}                                                 & \multicolumn{2}{c|}{Mean(SD):1.54(0.27)}                                                 \\ \hline
\multirow{2}{*}{\textbf{20$^0$}}     & 1.11                     & 2.20                                    & 1.10                                    & 2.13                                   \\ \cline{2-5} 
                                 & \multicolumn{2}{c|}{Mean(SD): 1.54(0.31)}                                                 & \multicolumn{2}{c|}{Mean(SD): 1.58(0.32)}                                                 \\ \hline
\multirow{2}{*}{\textbf{30$^0$}}     & 1.12              & 2.10                                    & 1.00                                    & 2.00                                    \\ \cline{2-5} 
                                 & \multicolumn{2}{l|}{Mean(SD): 1.52(0.28)}                                                 & \multicolumn{2}{l|}{Mean(SD): 1.54(0.30)}                                                 \\ \hline
\end{tabular}
\end{table}

During the trails, the cutting blade moved along with the vertical ram of the UTM (at the rate of 1.5 mm/min) pierced the specimen center, and cut the specimen peduncle into two halves. This simulates the peduncle cutting happening at 15 mm from the top surface of strawberry using a parallel cutting mechanism. While the blade was interacting with the stem, the load cell values were recorded by the connected PC and were used to identify the peak force required for cutting each specimen. The force recordings are presented in section~\ref{sec:results}.

\section{Results and Discussion}
\label{sec:results}

\subsection{Gripping force study}
\begin{figure}[tb!]
\centerline{\includegraphics[scale=0.63]{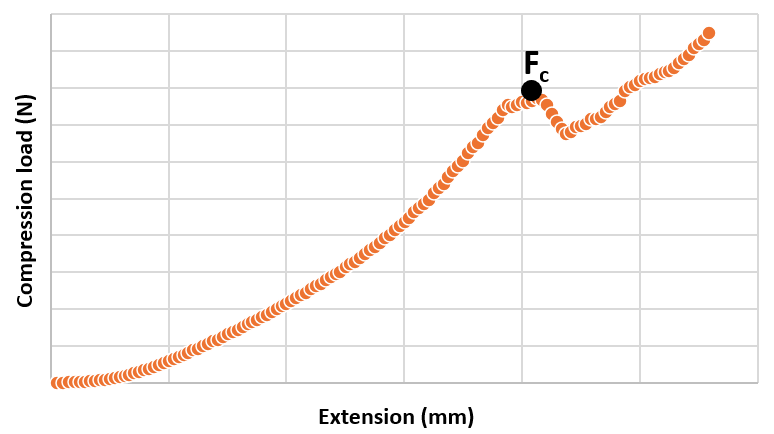}}
\caption{Force profile recorded for the specimens while applying compression load by the extending UTM ram. The final squeezing force is not presented here as the peak force (F$_{C}$) before sudden drop is the point of interest}
\label{fig4}
\end{figure}

\begin{figure}[tb!]
\centerline{\includegraphics[scale=0.31]{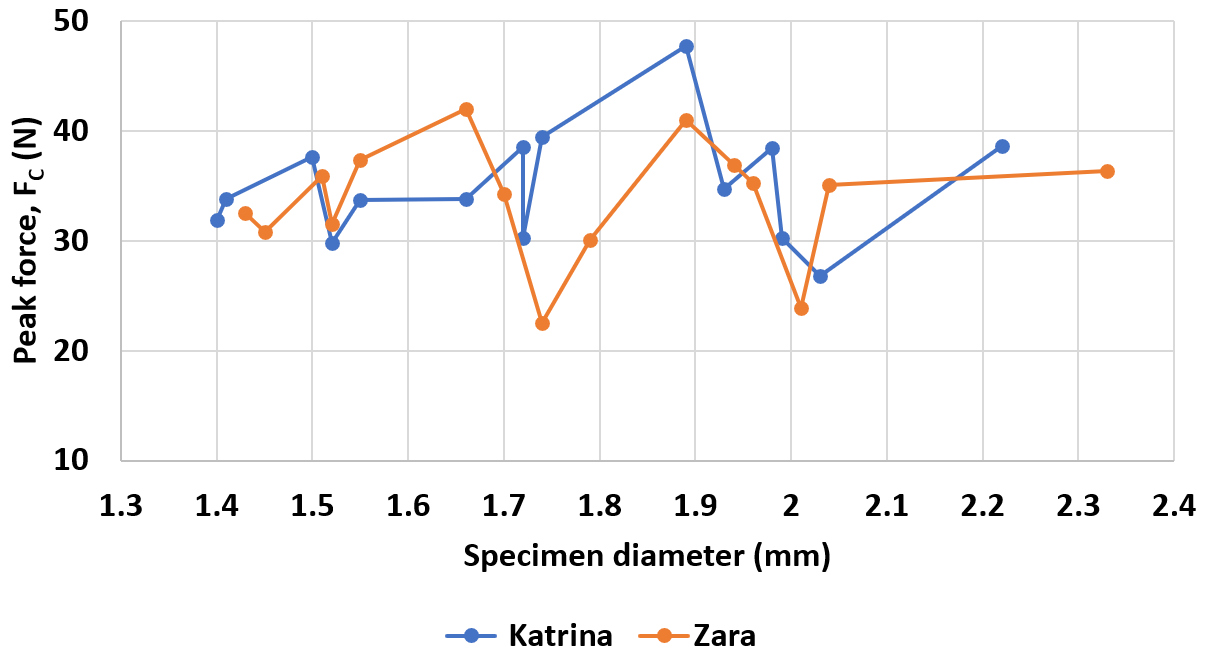}}
\caption{Variation of F$_{C}$ with respect to the specimen diameter}
\label{gripVsD}
\end{figure}

%\subsubsection{Gripping force test}
During the experimental trials, all tested specimens showed a common force profile under compression load which is shown in the figure \ref{fig4}. While applying compression load (by the extending UTM ram) to the specimen, there was a gradual increase in the resistive force to a certain point, from then it showed a sudden drop. After then, the specimen got squeezed completely on further application of compression load. It has been noticed that, after the drop in the resistive force, the specimen went to permanent deformation before getting completely squeezed. This peak force before the drop (F$_{C}$), is considered as the point of interest and is recorded for all specimens against their respective diameter (figure~\ref{gripVsD}). The trend of this force (F$_{C}$) can be studied to limit the gripping force on the peduncle such that there is a lesser chance of squeezing the peduncle during the gripping action. %The squeezing or crushing of the peduncle during the gripping action can result the detached strawberry falling off from the grip during the harvesting process.

\begin{figure}[tb!]
\centerline{\includegraphics[scale=0.575]{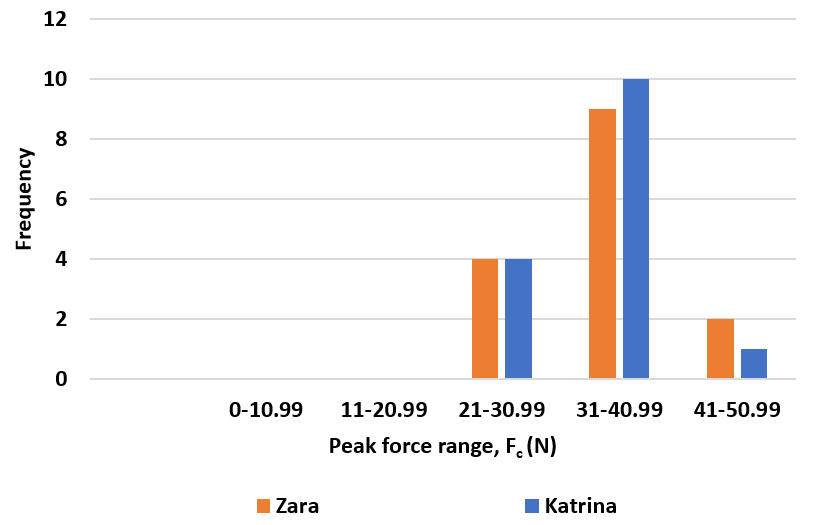}}
\caption{Frequency distribution of F$_{C}$ measured for all specimens}
\label{Result10}
\end{figure}

Considering the peak forces, presented in the figure  \ref{gripVsD}, we can determine that the lowest of these peak forces (F$_{C}$) recorded is about 26.83 N and 22.53 N for Katrina and Zara respectively. This means that at these lowest values of compression force (analogous to gripping force), the respective test specimen went into permanent deformation before squeezing. Hence if we allocate a factor of safety of 2 to the lowest of these two values of forces (26.83 N and 22.53 N), the gripping force should be limited to around 10 N. Assuming that for a two jaw parallel gripper, if we use a soft material lining on the gripping surface with a coefficient of friction of 0.3, then under static condition (see equation \ref{eqn1}, a = 0 m/s$^2$), the end effector will be able to handle a mass of 300 grams. Assuming that the mass of the strawberry to be gripped in the end effector is about 50 grams and keeping other parameters the same, under dynamic conditions, the end effector would be able to handle this 50 gram strawberry for a manipulator acceleration of up to 50 m/s$^2$. 
\begin{equation}
F_g=\frac{m.(g+a).S}{\mu}
\label{eqn1}
\end{equation}
where; 'F$_g$' is the net gripping force (N), 'm' is the mass to be handled (Kg), 'g' is the acceleration due to gravity (m/s$^2$), '$\mu$' is the co-efficient of friction, and 'S' is factor of safety\cite{Grippingforce}. 

This computed load carrying capability at 10 N is more than enough for an end effector to handle a ripe strawberry in static and dynamic conditions of a harvesting cycle. It is to be noted that, in dynamic conditions, for the same gripping force, the loading capacity can be increased by regulating the manipulator acceleration.

The frequency distribution of the peak forces (F$_{C}$) of the tested specimens is shown in figure~\ref{Result10}. The majority of the forces (F$_{C}$) fall in the range of 31.00 N to 40.99 N for both the varieties, when the diameter of the specimens shares a close mean of 1.75 mm and 1.76 mm (with a standard deviation of 0.24 and 0.25) for Katrina and Zara respectively. Since the computed compression/gripping force limit (10 N) is well below this range, it can be concluded that 10 N would be a safe force that can be applied to the peduncle for an effective gripping. 
\begin{figure}[h!]
\centerline{\includegraphics[scale=0.55]{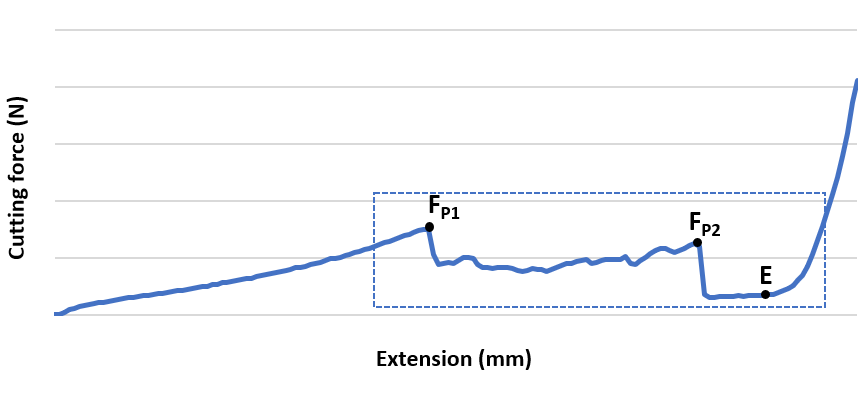}}
\caption{Force profile recorded for the specimens during cutting test}
\label{fig6}
\end{figure}

\begin{figure*}[h!]
\centerline{\includegraphics[scale=0.63]{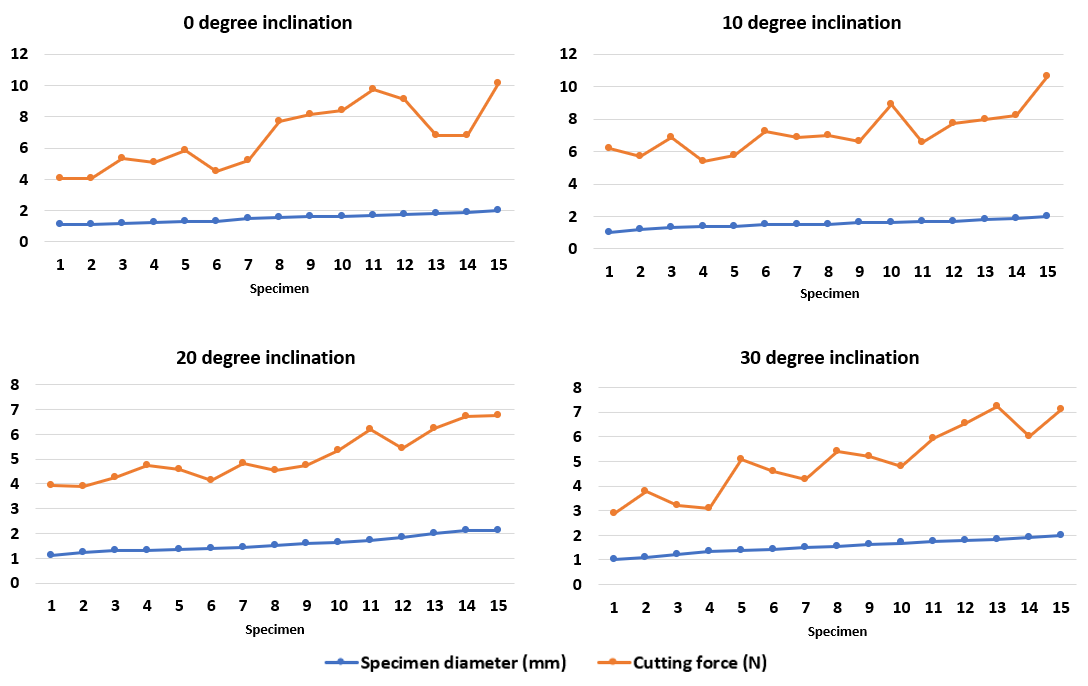}}
\caption{Variation of F$_P$ with respect to specimen diameter at test cutting orientations (Katrina)}
\label{figCK}
\end{figure*}

\begin{figure*}[h!]
\centerline{\includegraphics[scale=0.63]{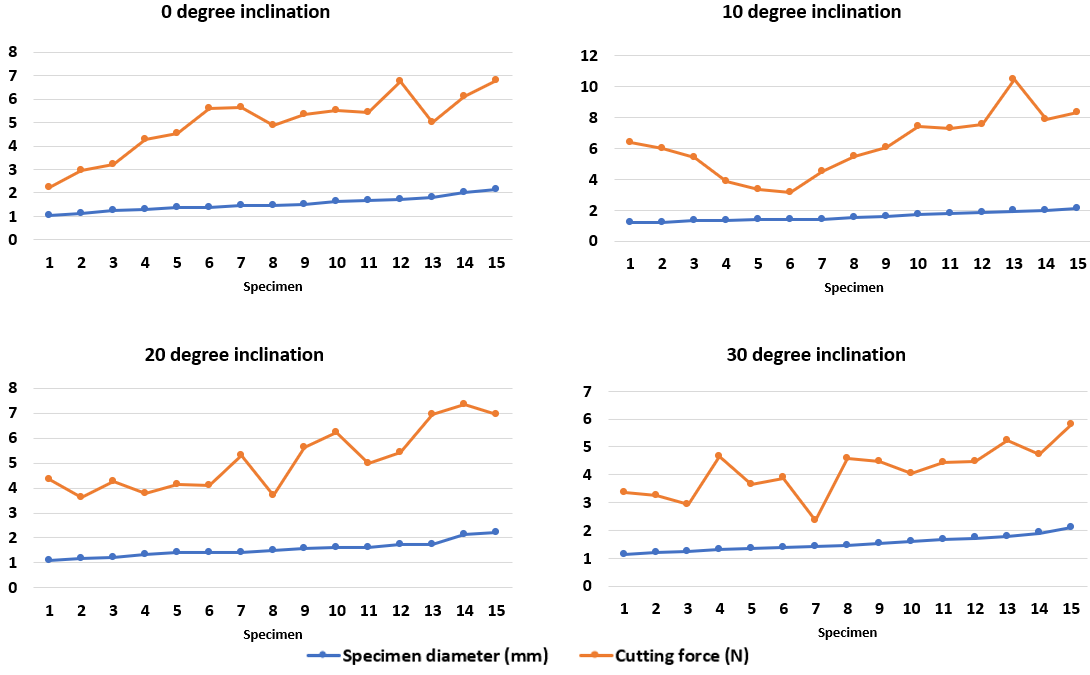}}
\caption{Variation of F$_P$ with respect to specimen diameter at test cutting orientations (Zara)}
\label{figCZ}
\end{figure*}

\subsection{Cutting force study}
While analyzing the force values recorded during the cutting trials, again a common force profile has been noticed for all the tested specimens. A sample force profile is shown in the figure \ref{fig6}. In the profile, there is an increase in force value during the cutting action but with two sudden drops after two force peaks (F$_{P1}$ and F$_{P2}$). After the second peak force, there is a flat profile and followed by a sharp rise in force after a point (E). This sharp increase happens when the blade touches the peduncle supports after cutting the peduncle off. So from the force profile for each specimen, the maximum of the two peak forces (F$_{P1}$ or F$_{P2}$) is taken as the peak cutting force (F$_P$) required for that specimen. The peak cutting forces recorded for all specimens against their diameter and at different orientations are shown in the figure \ref{figCK} and \ref{figCZ}.

From the force values recorded at different cutting orientations, it has been noticed that the mean cutting force shows a relative lowest value at 30$^0$ orientation compared to other studied orientations. These mean cutting force values (F$_{PM}$) with respect to the mean diameter (D$_M$) of the tested specimens at different orientations are presented in table \ref{table2}. Also, it is seen that at 10$^0$ orientation, the mean cutting force (F$_{PM}$) shows a relatively higher value. These variations are found to be true for both the strawberry varieties studied.

\begin{table}[tb!]
 \caption{Variation of  F$_{PM}$ with D$_M$ of specimens at test cutting orientations. Standard deviations are presented in brackets}
       \begin{tabular}{|c|c|c|c|c|}
       \hline
       \textbf{Orientation} & \textbf{D$_M$ (mm)} & \textbf{F$_{PM}$ (N)} & \textbf{D$_M$ (mm)} & \textbf{F$_{PM}$ (N)} \\
        \textbf{of cut} & \textbf{[Zara]} &\textbf{[Zara]} & \textbf{[Katrina]} & \textbf{[Katrina]} \\
        \hline
        0$^0$ & 1.53(0.31) & 4.95(1.33) & 1.51(0.29) & 6.71(2.05) \\ \hline
        10$^0$ & 1.59(0.30) & 6.22(2.03) & 1.54(0.27) & 7.17(1.36) \\ \hline
        20$^0$ & 1.54(0.31) & 5.11(1.27) & 1.58(0.32) & 5.08(0.98) \\ \hline
        30$^0$ & 1.52(0.28) & 4.13(0.90) & 1.54(0.30) & 5.00(1.40) \\ \hline
    \end{tabular}
    \label{table2}
\end{table}
At 30$^0$ cutting orientation, the maximum of the peak cutting force (F$_P$) recorded is about 7.20 N for Katrina, and 5.80 N for Zara. And hence, with a factor of safety 2, the cutting force requirement can be approximated to 15 N which could be considered sufficient to cut the strawberry peduncle at 30$^0$ orientation. Also, this force would be sufficient to handle other cutting orientations studied. It is to be noted that this cutting force is measured at a blade speed of 1.5 mm/min, so at higher blade speed, the required force will be lesser than 15 N.

\section{Conclusions}

This work reports the studies conducted to estimate a few key parameters which can help in the design of strawberry harvesting EEs targeting the peduncle. The studied parameters include, the limit of gripping force that can be applied on the peduncle without crushing them, and the force required to cut peduncle. These parameters are estimated for two varieties of strawberry namely Katrina and Zara.
%From the field studies, the highest mass of the ripe strawberry (M$E$) studied is about 30.00 grams for Katrina and 32.54 grams for Zara and the lowest in the order of 7.00 grams and 6.69 grams. When examining the peduncle diameter, the highest and lowest value is in the order of 2.76 mm to 1.00 mm (Katrina) and 2.85 mm to 1.20 mm (Zara) respectively. 
The experimental studies have revealed that a safe gripping force of 10 N can be applied to the strawberry peduncle. And this gripping force is sufficient for a two jaw parallel gripper to retain a strawberry with a mass of 50 grams at a manipulation acceleration up to 50 m/s$^2$ and without squeezing the peduncle. With regards to the peduncle cutting force, it is shown that a blade with a wedge angle of 16.6$^0$ can cut the strawberry peduncle with a force of 15 N at an orientation of of 30$^0$. The cutting forces at an orientation of 30$^0$ have been studied as slightly lowest among the other cutting orientations of 0$^0$, 10$^0$, 20$^0$. This cutting force estimation will be valid while using a similar standard blade profile in an end effector with a parallel cutting mechanism.

We will make use of these parameters in designing a novel end effector for strawberry harvesting, and we present them here in the hope that they will be helpful to other researchers who are engaged in developing end effectors for harvesting strawberries.  

\section{Limitation of the study}

This study has only used specimens from two varieties of strawberries, and it is an untested assumption that the estimated figures and trends can be a representation of other varieties of strawberries within close tolerances. Since our proposed end-effector design calls only for blade orientation below 30$^0$, we have not considered angles above 30$^0$ in the cutting force study. Hence the cutting force trend for greater blade orientations remains unknown.

%\section*{Acknowledgment}

%The preferred spelling of the word ``acknowledgment'' in America is without 
%an ``e'' after the ``g''. Avoid the stilted expression ``one of us (R. B. 
%G.) thanks $\ldots$''. Instead, try ``R. B. G. thanks$\ldots$''. Put sponsor 
%acknowledgments in the unnumbered footnote on the first page.

\printbibliography
\end{document}